\documentclass[11pt, letterpaper, shortlabels]{berkeley}

\usepackage{amsmath,amsfonts,bm}

\def\eqref#1{equation~\ref{#1}}

\def\1{\bm{1}}

\DeclareMathAlphabet{\mathsfit}{\encodingdefault}{\sfdefault}{m}{sl}
\SetMathAlphabet{\mathsfit}{bold}{\encodingdefault}{\sfdefault}{bx}{n}

\usepackage{authblk}

\usepackage{amsmath, amsthm, amssymb, mathtools, mathrsfs}
\usepackage{graphicx, booktabs, multirow, wrapfig, subcaption}
\usepackage{enumitem, array, url, microtype, dsfont, nicefrac}
\usepackage{tcolorbox}
\tcbuselibrary{skins, breakable, listings, theorems}
\usepackage{setspace, lipsum, fontawesome5}
\usepackage{natbib}
\usepackage{soul}

\tcbset{
  definitionbox/.style={
    colback=white,
    colframe=black,
    breakable,
    enhanced,
    boxrule=0.5pt,
    arc=1mm,
    left=4pt,
    right=4pt,
    top=4pt,
    bottom=4pt,
  },
}

\usepackage{algorithm}
\usepackage{algorithmic}

\definecolor{NDblue}{RGB}{12, 35, 64}
\definecolor{NDgold}{RGB}{174, 145, 66}
\definecolor{darkblue}{rgb}{0, 0, 0.5}
\definecolor{deepblue}{rgb}{0,0,0.5}
\definecolor{deepred}{rgb}{0.6,0,0}
\definecolor{deepgreen}{rgb}{0,0.5,0}
\definecolor{NDblue}{RGB}{12, 35, 64}
\definecolor{NDgold}{RGB}{174, 145, 66}
\definecolor{bestcell}{RGB}{220,255,220}
\definecolor{modelrow}{gray}{0.95}
\definecolor{promptgray}{RGB}{200,200,200}
\definecolor{promptblue}{RGB}{25,118,210}
\PassOptionsToPackage{table}{xcolor}  
\usepackage[table]{xcolor}  
\usepackage{hyperref}

\hypersetup{
    colorlinks = true,
    linkcolor = NDblue,
    citecolor = NDgold,
    urlcolor  = NDblue,
    filecolor = NDblue
}

\usepackage[capitalize,noabbrev]{cleveref}
\Crefformat{equation}{#2Eq.~(#1)#3}
\Crefformat{figure}{#2Figure~#1#3}
\Crefformat{assumption}{#2Assumption~#1#3}
\Crefname{assumption}{Assumption}{Assumptions}
\usepackage{crossreftools}
\usepackage{listings}
\lstdefinestyle{pythonstyle}{
    basicstyle=\ttfamily\footnotesize,
    language=Python,
    keywordstyle=\color{deepblue},
    stringstyle=\color{deepgreen},
    frame=single,
    showstringspaces=false,
}
\lstnewenvironment{python}[1][]{
    \lstset{style=pythonstyle,#1}
}{}

\newtcolorbox{promptbox}[2][]{%
    enhanced,
    unbreakable,
    before skip=2mm,
    after skip=2mm,
    colback=darkblue!5!white,
    colframe=darkblue,
    coltitle=white,
    boxrule=0.5mm,
    sharp corners,
    arc=5pt,
    attach boxed title to top center={yshift=-3mm},
    boxed title style={
        enhanced,
        colback=NDgold,
        colframe=darkblue,
        arc=5pt,
        outer arc=5pt,
        boxrule=0pt,
    },
    title={\faLightbulb[solid]\space #2},
    fonttitle=\bfseries\color{white},
    #1
}

\title{Better Datasets Start From RefineLab: Automatic Optimization for High-Quality Dataset Refinement}
\author[1]{Xiaonan Luo$^\ast$}
\author[1]{Yue Huang$^\ast$}
\author[2]{Ping He}
\author[1]{Xiangliang Zhang}

\affil[1]{University of Notre Dame}
\affil[2]{Vanderbilt University}
\correspondingauthor{xzhang33@nd.edu (Xiangliang Zhang). $^\ast$These authors contributed equally to this work.}

\begin{abstract}
High‑quality Question–Answer (QA) datasets are foundational for reliable Large Language Model (LLM) evaluation, yet even expert‑crafted datasets exhibit persistent gaps in domain coverage, misaligned difficulty distributions, and factual inconsistencies. 
The recent surge in generative model-powered datasets has compounded these quality challenges. In this work, we introduce \textbf{RefineLab}, the first LLM‑driven framework that automatically refines raw QA textual data into high-quality datasets under a controllable token‑budget constraint. 
RefineLab takes a set of target quality attributes (such as coverage and difficulty balance) as refinement objectives, and performs selective edits within a predefined token budget to ensure practicality and efficiency. In essence, RefineLab addresses a constrained optimization problem: improving the quality of QA samples as much as possible while respecting resource limitations. 
With a set of available refinement operations (e.g., rephrasing, distractor replacement), RefineLab takes as input the original dataset, a specified set of target quality dimensions, and a token budget, and determines which refinement operations should be applied to each QA sample. This process is guided by an assignment module that selects optimal refinement strategies to maximize overall dataset quality while adhering to the budget constraint. Experiments demonstrate that RefineLab consistently narrows divergence from expert datasets across coverage, difficulty alignment, factual fidelity, and distractor quality. RefineLab pioneers a scalable, customizable path to reproducible dataset design, with broad implications for LLM evaluation.

\end{abstract}
\begin{document}
\maketitle

\section{Introduction}\label{sec:intro}
Large language models (LLMs) have achieved remarkable success on a variety of natural language understanding and generation tasks, driving rapid progress in various areas \citep{zhao2023survey, guo2023what, qian2023chatdev}. To better evaluate and improve the capabilities of these models, central to this progress is the availability of high‑quality datasets \citep{wang2024mmluprorobustchallengingmultitask, lin2022truthfulqameasuringmodelsmimic, zellers2019hellaswag}. 
A high-quality evaluation dataset for LLMs is widely regarded as one that reliably reflects model capabilities, minimizes bias, and facilitates fair comparisons across models.   Although this notion is broadly accepted within the community, there is no formal standard that explicitly defines what constitutes a high-quality dataset, or quantifies how the quality of a dataset influences evaluation reliability. In this work, we seek to reduce this ambiguity by identifying and analyzing key dimensions of dataset quality that can be systematically enhanced through careful sample refinement, while accounting for practical constraints such as token-level editing budgets.

\begin{figure}
    \centering
    \includegraphics[width=\linewidth]{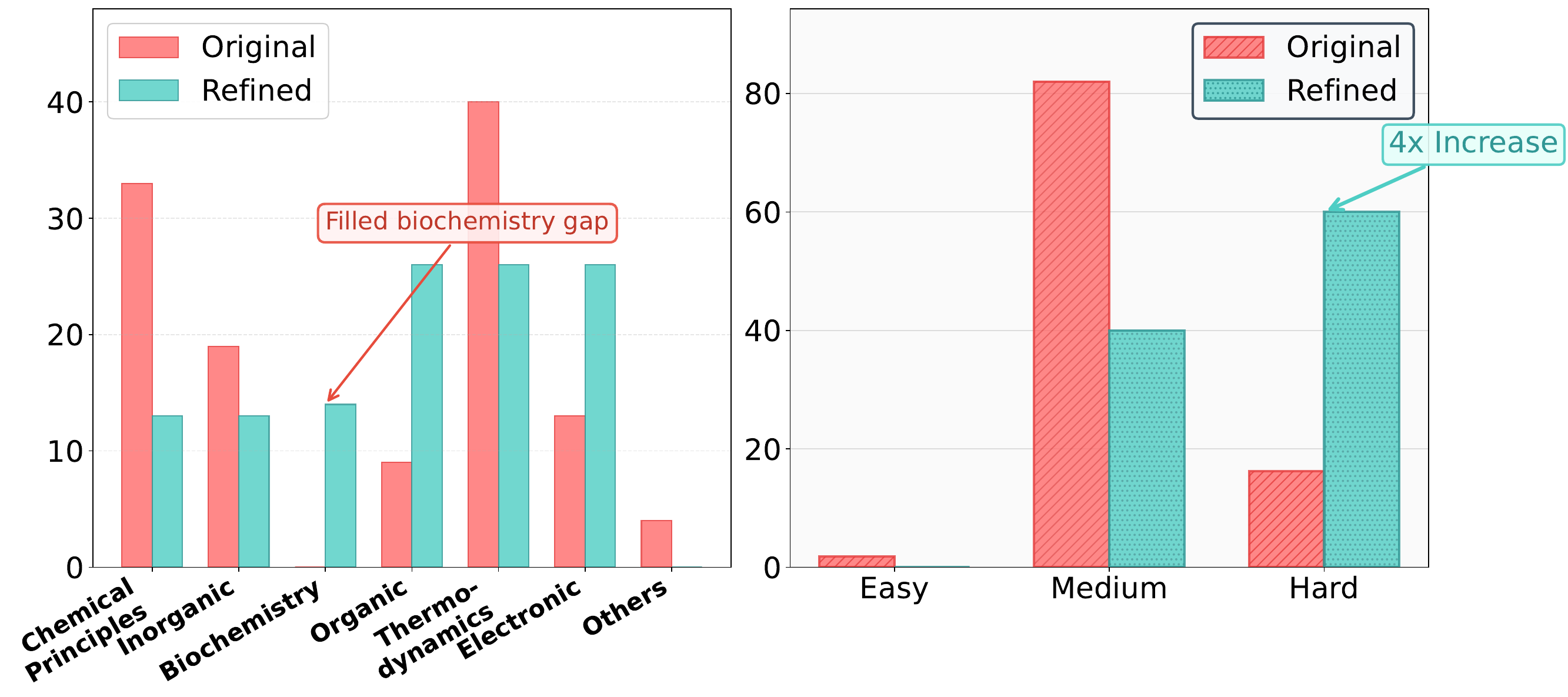}
    \caption{An example of topic coverage (left) and difficulty level distribution (right) in the College Chemistry subset of MMLU before (Original) and after refinement (Refined). Targets follow MIT’s undergraduate chemistry major curriculum and feature a stress-testing difficulty profile.}
    \label{fig:example}
\end{figure}

To ground our discussion, we take the widely adopted MMLU dataset \citep{hendrycks2021measuring} as an illustrative example. MMLU is often praised for its broad subject coverage and standardized multiple‑choice format, which makes it suitable for evaluating knowledge recall and reasoning across diverse domains. However, despite its popularity, MMLU exhibits several key limitations: \textbf{1) Topic coverage:} its empirical topic mix can diverge sharply from real-world evaluation standards, for example, in the College Chemistry subset shown in Figure~\ref{fig:example} (left), \textit{biochemistry} is completely absent, yet real‑world college curricula, including MIT, Stanford, and other world-class universities, always demand a certain portion representation on it, resulting in datasets that fail to adequately reflect the full range of domain‑specific skills and knowledge. \textbf{2) Imbalanced Difficulty Distribution:} its fixed difficulty profile cannot accommodate varying evaluation goals, whether stress‑testing advanced reasoning for math and science or supporting progressive skill diagnostics for language acquisition. As Figure~\ref{fig:example} (right) shows, the original College Chemistry subset is dominated by medium-level difficulty questions that lack deep reasoning. Under an example of \textit{stress-testing} target profiles, naive resampling can only remove excess medium items, drastically shrinking the dataset. Further refinement is needed to introduce new examples and maintain the effective dataset size while meeting specific difficulty requirements. \textbf{3) Factual and Logical Errors:} its factual or logical mistakes that can persist unchecked, introducing misleading artifacts into model evaluation (for example, \citep{wang2024mmluprorobustchallengingmultitask} pointed out that 6.5\% of all questions in MMLU contain errors). Such errors can compromise the validity of benchmarking results. These shortcomings highlight the necessity of refining existing datasets to ensure reliable model evaluation and to drive progress in LLM development and downstream applications.

While dataset refinement is essential, manual approaches are often tedious and inefficient. Recent work has introduced LLM-based methods for automatic dataset refinement, leveraging LLMs to synthesize new data with minimal human involvement~\citep{khan2025dataenvgym, li2025gendataagent, huang2025datagen, desalvo2025softsrvlearngeneratetargeted, li2025autobencher}. Although these approaches are promising, several challenges remain: 
\textbf{1) Lack of explicit control:} LLM-driven pipelines may unintentionally amplify coverage gaps or difficulty imbalances, as synthesized examples are often generated without fine-grained constraints. 
\textbf{2) Quality assurance limitations:} Without rigorous validation, LLM-generated data can introduce new factual or logical errors, complicating overall quality control. 
\textbf{3) Resource allocation dilemma:} Given that invoking high-capability models (e.g., GPT-4) for refinement incurs significant computational cost, efficiently allocating a limited budget of LLM calls across multiple refinement strategies to maximize dataset quality remains a challenging and largely unsolved problem.

To end this, in this paper, we introduce \textbf{RefineLab}, a framework for textual dataset refinement that automatically selects and applies targeted editing operations to improve dataset quality under token-level budget constraints. In RefineLab, dataset \emph{high quality} is operationalized via explicitly specified targets (e.g., desired topic distribution for coverage, difficulty levels, or factual accuracy). These targets can be defined either by users or inferred from real-world evaluation needs. Given such explicit requirements and a set of available refinement strategies, RefineLab employs an assignment module that selects the optimal refinement operations for each sample to maximize overall dataset quality, while respecting a global token-level budget constraint.
Based on the estimated cost and quality gain of each refinement option, the assignment problem is formulated as an Integer Linear Programming (ILP) optimization, which determines,  for each QA sample $(q_i,a_i)$, which refinement operations ($r_k$) to apply: $x_{ik}$ =1 if refinement operation  $r_k$  is assigned to sample  $(q_i,a_i)$, and  $x_{ik}$ = 0 otherwise. 
Overall, the objective is formalized as
\begin{equation}\label{eq:ILP}
\max_{x_{ik} \in \{0,1\}}  \sum_{i,k} x_{ik} \cdot \Delta_{ik}   \quad 
\text{s.t.} \quad   \sum_{i,k} x_{ik} \cdot c_{ik} \leq B
\end{equation}
where   \(\Delta_{ik}\) denotes the estimated quality gain from applying \(r_k\) to sample \((q_i, a_i)\),  \(c_{ik}\) represents the token cost of that refinement, and \(B\) is the total available token budget.
This formulation enables RefineLab to reason about trade-offs between different refinement operations and their associated costs, providing a principled way to enhance dataset quality under resource constraints.

In this work, we make the following key contributions:
\begin{itemize}
  \item We propose \textbf{RefineLab}, the first LLM-driven framework for dataset refinement that formulates the refinement selection process as an \textit{integer linear program} to optimize quality improvements under a token-level budget.
  \item We introduce a set of modular refinement operations designed to improve QA datasets along a set of explicit quality targets, including topic coverage, difficulty calibration, and factual consistency.
  \item We conduct comprehensive experiments on widely-used benchmarks such as MMLU, demonstrating that  RefineLab produces higher-quality QA datasets with improved topic coverage, difficulty balance, and overall utility for evaluating LLM performance.
\end{itemize}

\vspace{-5pt}
\section{Related Work}
\label{sec:related_work}
\paragraph{Synthetic Dataset Generation.}

LLMs have enabled scalable, automated construction of synthetic datasets via prompting. Self‑Instruct \citep{wang-etal-2023-self-instruct} bootstraps instruction–response pairs through iterative self‑prompting and employs minimal manual filtering to ensure basic correctness. DataGen \citep{huang2025datagen} introduces a modular framework using attribute‑guided prompts and automated consistency checks to generate diverse text, supplemented by lightweight heuristic filters for domain validity. AutoBencher \citep{li2025autobencher} advances a declarative dataset construction paradigm, allowing users to specify high‑level dataset schemas from which LLMs generate evaluation items under explicit quality constraints. However, these methods focus on data synthesis that is typically guided by high-level prompts with limited control over sample-level quality dimensions. In contrast, RefineLab targets systematic dataset improvement by selecting editing operations through constrained optimization, enabling fine-grained, budget-aware refinement of real datasets.

\paragraph{Evaluation on Dataset Quality.}

High‑quality datasets are fundamental to effective LLMs training and reliable evaluation. Recent works emphasize that current dataset flaws, including distributional bias, uncalibrated factual errors, lead to misleading performance estimates or inefficient model training \citep{reuel2024betterbench, yu2023large, xie2023data, miranda2025scalediversitycoefficientdata, iskander-etal-2024-quality}. As a result, ensuring high‑quality datasets becomes essential. Recent efforts have focused on designing metrics that capture multiple, complementary aspects of data validity. For example, factual integrity is typically assessed through error‑detection and correction pipelines that retrieve external evidence to validate item correctness, achieving high correction rates in benchmark evaluations \citep{fatahi-bayat-etal-2023-fleek}. In multiple‑choice settings, distractor quality is measured along dimensions of plausibility and variety: automatic metrics for incorrectness and semantic similarity provide quantitative assessments \citep{raina-etal-2023-assessing}, student‑choice‑prediction models estimate diagnostic power by modeling real‑user error patterns \citep{lee-etal-2025-generating-plausible}, and synthesize best practices for distractor generation and evaluation \citep{alhazmi-etal-2024-distractor}.
RefineLab is a flexible and extensible framework that can incorporate a wide range of quality metrics along with their corresponding refinement operations,  enabling targeted improvements in a controllable and resource-aware manner.

\section{The RefineLab Framework}
\label{sec:method}
\subsection{Notations and Formulation}

Given a raw textual dataset with questions \(q\) and corresponding answers \(a\), denoted as  $ D = \{(q_i, a_i)\}_{i=1}^N$, and a set of specified quality targets \(\mathcal{T} = \{t_1, t_2, \ldots, t_k\}\), the goal of  RefineLab is to transform \(D\) into a refined dataset \(D'\) by applying a set of selected refinement operations. Formally, we define a refinement process 
\[
\mathcal{R} : (D, \mathcal{T}) \;\mapsto\; D',
\]
where \(\mathcal{R} = \{r_1, r_2, \ldots, r_k\}\) is the set of available refinement operators applied selectively to each sample \((q_i, a_i)\). The assignment of refinement operations is determined by solving the ILP problem described in Eq.~(\ref{eq:ILP}), such that the resulting dataset \(D'\) better satisfies the specified quality targets \(\mathcal{T}\) while adhering to the token-level budget constraint.

Each quality target \( t_k \in \mathcal{T} \) represents a desired property of the dataset such as a target distribution over \textit{topics}, \textit{difficulty levels}, or \textit{factual consistency}. These targets can be either explicitly provided by users (e.g., for custom evaluation settings) or automatically derived from real-world benchmarks or downstream task requirements.
For simplicity, we assume that each refinement operator \( r_k \in \mathcal{R} \) is designed to address a specific quality target \( t_k \in \mathcal{T} \). That is, there exists a one-to-one mapping between refinement operations and quality targets, where each \( r_k \) is specialized to improve the dataset with respect to the corresponding \( t_k \). Therefore, the quality gain \(\Delta_{ik}\) in Eq.~(\ref{eq:ILP}), which represents the improvement from applying refinement operation \(r_k\) to sample \((q_i, a_i)\), is defined with respect to the corresponding target \(t_k\), reflecting how well the refined sample aligns with the desired quality dimension addressed by \(r_k\).

\begin{figure*}[t]
  \centering
  \includegraphics[width=\linewidth]{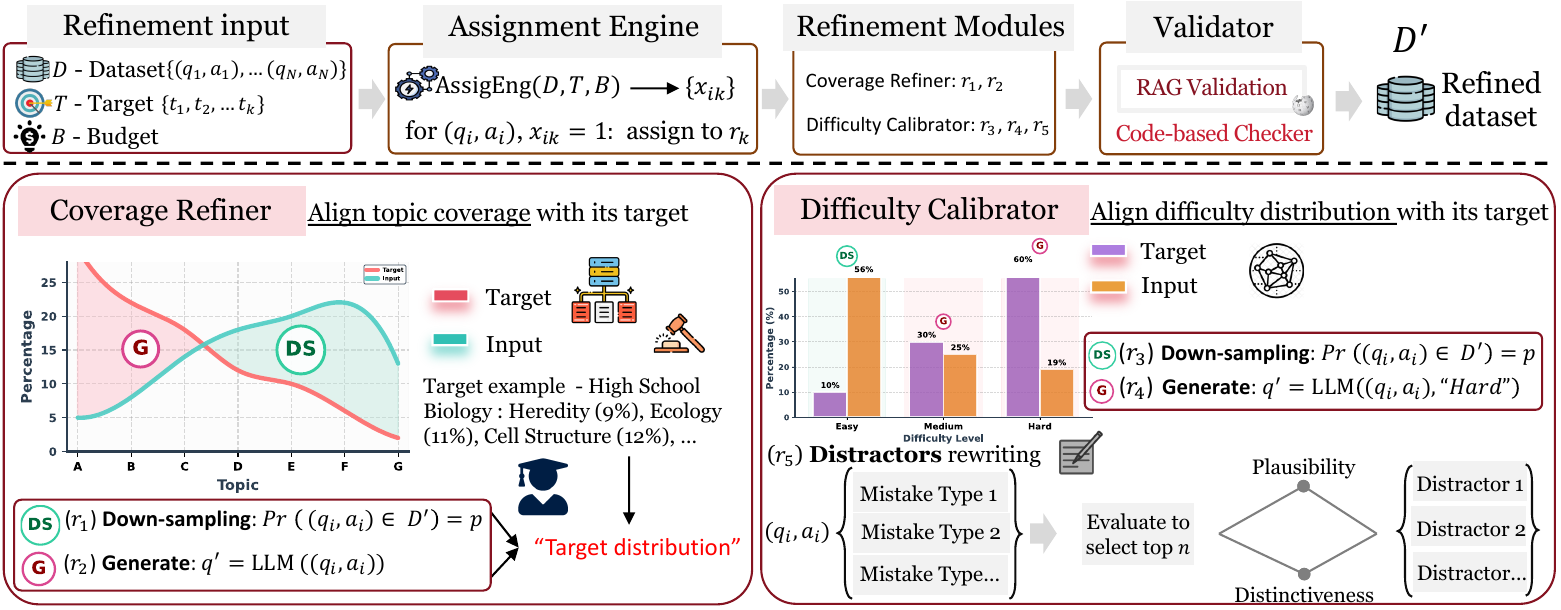}
  \caption{Top: the framework of RefineLab. Bottom: the Refinement Operations.}
  \label{fig:framework}
\end{figure*}

The RefineLab framework is shown in the top part of Figure \ref{fig:framework}. We next introduce the core components of the refinement process, including the design of refinement operations,  the mechanism by which the ILP is solved by the assignment engine to determine optimal refinement decisions, and the data sample validation process.

\subsection{Data Refinement Operators}

\subsubsection{Coverage Refiner (\(r_1, r_2\)).}
Datasets collected without reference to authorized guidelines often exhibit uneven representation of topics or skills. For example, in the \textit{College Chemistry} subset of MMLU, fewer than 1\% data cover organic chemistry, despite organic chemistry typically comprising over 10\% of a standard college curriculum. Such under‑representation can inflate model ability on over‑represented topics while obscuring weaknesses in critical domains. To mitigate this issue, the \textit{Coverage Refiner} realigns the dataset’s empirical domain–skill distribution to a specified target vector \(t\). This distribution vector could be provided 
by the user or drawn from a set of empirically validated and authorized human‑examination guidelines, e.g., AP curriculum standards, grade‑level learning objectives, to ensure that each domain category is represented according to specification. 

 The distribution alignment can be achieved by \textit{increasing} the number of samples in underrepresented categories and \textit{reducing} those in overrepresented categories, each incurring different costs. While the \textit{assignment engine} determines which action should be applied to each QA sample, the \textit{Coverage Refiner} operates on the assigned sample \((q,a)\) with two possible actions: 
(1) \textbf{Removal (\(r_1\))},  if fewer samples from the category of \((q,a)\) are needed, it drops \((q,a)\) from the dataset. The probability of removing a sample \((q,a)\) is determined by the degree of \textit{overrepresentation} of its category relative to the target distribution \(t\); 
(2)  \textbf{Expansion (\(r_2\))},  if more samples from the category of \((q,a)\) are needed, it generates additional instances by rephrasing \((q,a)\). Specifically, the operator takes in \((q,a)\) and a target topic (\(t^{\text{topic}}\)), prompts an LLM to produce lexically varied QA pairs that preserve the original question’s topic focus and correctness. The generation hint includes the original QA pair, a brief instruction to maintain fidelity to the source topic and reasoning logic, and optionally a target lexical style or variation constraint. Multiple paraphrased versions are sampled, followed by lightweight filtering to eliminate degenerate or duplicate outputs. The detailed prompt template can be found in the technical appendix B.

\subsubsection{Difficulty Calibrator (\(r_3, r_4, r_5\)).}
Real‑world datasets frequently exhibit skewed difficulty distributions. For example, prior work has noted that most questions in MMLU emphasize knowledge recall over reasoning, highlighting the need to recalibrate difficulty levels and tailor the question mix to specific research tasks and user requirements. To address such concerns, we design the \textit{Difficulty Calibrator} module to adjust a dataset’s difficulty distribution via efficient pairwise comparisons against a curated seed set. This allows us to infer the relative difficulty of each question and subsequently align the overall difficulty distribution with a specified target, e.g., the desired percentage of samples in \textit{easy}, \textit{medium}, and \textit{difficult} categories. 

Specifically, calibration can be achieved in three ways:  
(1) \textbf{Removal (\(r_3\))},  dropping samples from difficulty levels that are overrepresented relative to the target distribution;  
(2) \textbf{Generation (\(r_4\))},   synthesizing new questions to increase the proportion of underrepresented difficulty levels;  and
(3) \textbf{Distractor Rewriting (\(r_5\))},   modifying incorrect answer choices to adjust the cognitive load and thereby shift a question's difficulty level without changing its topic or factual correctness.
Each of these approaches incurs a different refinement cost, but all contribute to enriching the refined dataset by improving the balance of difficulty levels.



\textit{Generation (\(r_4\))} creates new QA samples at a desired difficulty level, involves prompt conditioning the LLM with explicit difficulty specifications \((t^{\text{diff}})\), a domain anchor (e.g., “college chemistry - thermodynamics”), and a reasoning indicator calibrated to the target level. For example, a \textit{hard} question may require multi-step reasoning or distractor designs that encode subtle misconceptions. The prompt template can be found in the technical appendix B.

\textit{Distractor Rewriting (\(r_5\))} is a cost-effective approach for adjusting the difficulty of multiple-choice questions. It modifies one or a few distractors while keeping the correct option intact. This method is generally less expensive than generating completely new questions, as it focuses on improving the quality of existing distractors rather than creating entirely new samples. It takes in a QA pair  \((q,a)\) and a target difficulty level \((t^{\text{diff}})\), then undergoes the following process to adjust the difficulty.
The process begins by prompting an LLM to identify the set of relevant mistake types for the QA sample, and to generate candidate distractors that exemplify each specific mistake type. These candidates are scored based on both their plausibility (i.e., how likely they are to be a reasonable distractor), measured by a confidence score given by the LLM generator as a plausible answer, and their typological distinctiveness (i.e., how different they are from the correct answer), computed via embedding-based cosine distance from the correct answer. The top \(n\) distractors are selected, ensuring that each mistake category is represented at least once in the final set of distractors, enriching the question’s difficulty without introducing significant overhead. Conversely, reducing difficulty can be achieved by removing or simplifying distractors with high semantic proximity to the correct answer or low plausibility. The detailed prompt template can be found in the appendix B.

To ensure that generated questions align with the intended difficulty levels, each new sample is compared against a curated seed set using pairwise Elo scoring \citep{elo1978rating}. First, exemplars with fixed difficulty ratings are sampled from the nearest semantic cluster of the generated question. An LLM is prompted to judge which sample is more difficult in each pairwise comparison. Based on the binary outcomes, the sample's provisional Elo score \(e_i\) is updated iteratively by 
\[
  e_i \leftarrow e_i + K_{\mathrm{Elo}}\bigl(r_{ij}-p_{ij}\bigr),
  \quad
  p_{ij} = \bigl(1 + 10^{(e_{\mathrm{ex}} - e_i)/\eta}\bigr)^{-1},
\]
where \(e_{\mathrm{ex}}\) is the exemplar’s Elo rating, and \(K_{\mathrm{Elo}}, \eta\) are hyper-parameters. After convergence, the final score \(e_i\) is mapped to a discrete difficulty band using pre-defined thresholds, and only those samples whose estimated difficulty matches the target are retained. The pairwise comparison prompt template is provided in the appendix B.

To determine whether a given sample should undergo \textit{Distractor Rewriting} or full \textit{Generation}, the system considers both the difficulty gap and token-level cost. If the difficulty change required (e.g., moving from easy to hard) exceeds what is achievable through distractor modification alone, generation is preferred despite higher cost. Otherwise, rewriting is selected for its cost efficiency. This decision is encoded in the \textit{assignment engine} based on estimated difficulty quality gain and refinement cost. 

\subsection{Assignment Engine}
As shown in Algorithm~\ref{alg:assignment_engine}, the main objective of the \textit{Assignment Engine} is to solve the ILP task and determine, for each QA pair \((q_i, a_i)\), the optimal refinement operation \(x_{ik}\in\{0,1\}\), along with the corresponding refinement targets: \(t^{\text{topic}}_i\) for coverage expansion and \((t_i^{\text{diff}})\) for difficulty calibration. These targets are sampled from underrepresented categories proportional to their mass of gaps between the empirical and target distribution. As shown in Eq.~(\ref{eq:ILP}), the ILP involves estimating \(\Delta_{ik}\), the quality gain from applying \(r_k\) to  \((q_i, a_i)\), and the token cost \(c_{ik}\) associated with that refinement.  


 The \textit{Assignment Engine} estimates \(\Delta_{ik}\) and \(c_{ik}\) in two steps using representative subsets. First, for each refinement operation \(r_k\) and candidate target \(z\) (e.g., a specific topic or difficulty level), we apply \(r_k\) to a pilot batch \(\mathcal{B}_{k,z}\) of QA samples conditioned on target \(z\). The average improvement in the relevant quality metric is recorded as:
\[
\Delta_{k,z} \approx \frac{1}{|\mathcal{B}_{k,z}|} \sum_{(q_j, a_j) \in \mathcal{B}_{k,z}} \Delta_{\text{metric}}(q_j, a_j, r_k, z),
\]
where \(\Delta_{\text{metric}}(\cdot)\) denotes the target-specific quality gain.

Similarly, the average token cost for invoking \(r_k\) under target \(z\) is estimated by:
\[
c_{k,z} \approx \frac{1}{|\mathcal{B}_{k,z}|} \sum_{(q_j, a_j) \in \mathcal{B}_{k,z}} \text{Token}(q_j, a_j, r_k, z).
\]

Then, for each QA pair \((q_i, a_i)\), the assignment engine sets \(\Delta_{ik} \leftarrow \Delta_{k,z_i}\) and \(c_{ik} \leftarrow c_{k,z_i}\), where \(z_i\) is the sampled \(t^{\text{topic}}_i\) or \(t_i^{\text{diff}}\), depending on the type of refinement.
This batched estimation scheme enables the ILP to reason about cost–benefit trade-offs efficiently, without the overhead of probing every sample individually. To mitigate the computational cost of solving the ILP at scale, we instead solve its LP relaxation to obtain fractional assignments \( \tilde{x}_{ik} \in [0,1] \), followed by a rounding step that converts them to binary \( x_{ik} \) while preserving the budget constraint.

\begin{algorithm}[tb]
\caption{Assignment Engine} \small
\label{alg:assignment_engine}
\textbf{Input}: dataset $D = \{(q_i, a_i)\}_{i=1}^N$; operations $\{r_k\}_{k=1}^K$;\\
target topic and difficulty distributions $\mathcal{T}$; budget $B$\\
\textbf{Output}: $\{x_{ik}\}$, $\{t^{\text{topic}}_i\}$, $\{t_i^{\text{diff}}\}$
\begin{algorithmic}[1]
\STATE Identify underrepresented bins in $\mathcal{T}$, e.g., topics \(\mathcal{T}_{\text{topic}}^u\)    and difficulties \(\mathcal{T}_{\text{diff}}^u \)  
\FOR{each $(q_i, a_i) \in D$}
  \STATE Sample $t^{\text{topic}}_i \sim \mathcal{T}_{\text{topic}}^u$,\quad $t_i^{\text{diff}} \sim \mathcal{T}_{\text{diff}}^u$
\ENDFOR
\FOR{each $r_k$ and target $z \in \mathcal{T} \cup \mathcal{D}$}
  \STATE Apply $r_k$ on pilot batch $\mathcal{B}_{k,z}$ to estimate $\Delta_{k,z}$, $c_{k,z}$
\ENDFOR
\FOR{each $(q_i, a_i)$ and operation $r_k$}
  \STATE Set $\Delta_{ik} \leftarrow \Delta_{k,z_i}$,\quad $c_{ik} \leftarrow c_{k,z_i}$
\ENDFOR
\STATE Solve ILP:
\[
\max_{x_{ik} \in \{0,1\}} \sum_{i,k} \Delta_{ik} x_{ik} \quad \text{s.t.} \quad \sum_{i,k} c_{ik} x_{ik} \le B
\]
\end{algorithmic}
\end{algorithm}

\subsection{Validator}

Previous studies have uncovered factual errors in widely used datasets. For example, the GSM8K dataset contains a proportion of incorrect solution labels, and some sociology sections of MMLU also include historically inaccurate statements. To mitigate these risks, the \textit{Validator} employs two complementary pipelines tailored to question type. 

For mathematical or multi‑step reasoning items, we use PoT prompting \citep{chen2023programthoughtspromptingdisentangling}: for each question–answer pair \((q_i,a_i)\), we prompt the LLM to produce a derivation trace $tri_i$, then apply a code-based checker \(\mathcal{J}\) such that \(\mathcal{J}(tri_i) = \mathrm{True}\) indicates procedural correctness. For factual or knowledge‑based questions, we extract a keyword set \(w_i\), retrieve relevant passages \(Rp(w_i)\) from Wikipedia, and feed \(\{q_i,a_i,Rp(w_i)\}\) into a verification prompt that asks the LLM to detect and correct any discrepancies. By combining PoT for numerical rigor with RAG \citep{lewis2021retrievalaugmentedgenerationknowledgeintensivenlp} for factual grounding, RefineLab achieves high integrity across both reasoning and factual domains. The detailed PoT and RAG validation prompt templates are provided in Appendix B.

\section{Experiments}

To evaluate  RefineLab, we conduct extensive experiments to answer the following evaluation questions (EQ):

\begin{itemize} 
\item \textbf{EQ1:} How RefineLab improves dataset quality across various dimensions;
\item \textbf{EQ2:} How RefineLab balances cost and quality;
\item \textbf{EQ3:} How   refined datasets affect  LLM evaluation;  
\item \textbf{EQ4:} How refined datasets help LLM fine-tuning.
\end{itemize}

\subsection{Experiments Setup}

\begin{table*}[ht]
  \centering
  \scriptsize
  \begin{tabular}{l
                  cc  
                  cc  
                  cc  
                  c   
                  c}  
    \toprule
    \textbf{Dataset}
      & \multicolumn{2}{c}{\textbf{Coverage (JSD)$\downarrow$}}
      & \multicolumn{2}{c}{\textbf{Difficulty (JSD)$\downarrow$}}
      & \multicolumn{2}{c}{\textbf{Distractor Diversity$\uparrow$}}
      & \textbf{Correction Ratio (\%)$\uparrow$}
      & \textbf{Error Rate (\%)$\downarrow$} \\
    \cmidrule(r){2-3}\cmidrule(r){4-5}\cmidrule(r){6-7}\cmidrule(r){8-8}\cmidrule(r){9-9}
      & Orig. & Refined 
      & Orig. & Refined 
      & Orig. & Refined 
      & Refined 
      & Refined \\
    \midrule
    MMLU       & 0.046 & 0.001 & 0.142 & 0.005 & 0.362 & 0.585 & 90.0 & 2.3 \\
    RACE       & 0.124 & 0.010 & 0.143 & 0.042 & 0.416 & 0.597 & 95.0 & 1.2 \\
    GSM8K      & 0.112 & 0.005 & 0.133 & 0.005 & --    & --    & 95.0 & 3.0 \\
    OpenbookQA & 0.300 & 0.012 & 0.320 & 0.090 & 0.239 & 0.456 & 94.4 & 1.0 \\
    HumanEval  & 0.280 & 0.002 & 0.131 & 0.021 & --    & --    & --   & 1.8 \\
    \bottomrule
  \end{tabular}
  \caption{Dataset quality evaluation before (Orig.) and after refinement (Refined). Correction Ratio and Error Rate are evaluated only on refined samples. (--) indicates not applicable, e.g., no errors in HumanEval or non‑MCQ datasets.}
  \label{tab:eq1_results}
\end{table*}

\paragraph{Dataset.} We evaluate RefineLab on the refinement of multiple widely-used datasets: \textbf{MMLU}: A multitask benchmark for evaluating knowledge recall and reasoning across 57 academic subjects. \textbf{RACE} \citep{lai-etal-2017-race}: A reading comprehension dataset sourced from English exams for Chinese middle and high school students. \textbf{GSM8K} \citep{cobbe2021trainingverifierssolvemath}: A grade school math word problem dataset requiring multi-step arithmetic and algebraic reasoning. \textbf{OpenbookQA} \citep{OpenBookQA2018}: A commonsense and science QA dataset based on an open-book of elementary-level science facts. \textbf{HumanEval} \citep{chen2021evaluating}: A code generation and completion benchmark featuring programming tasks with functional correctness tests.

\paragraph{Hyperparameters.} The hyperparameter setting includes default difficulty level targets \([0.0,0.4,0.6]\) (stress-testing profile with 40\% medium, 60\% hard) and coverage targets combining college curriculum, high‑school AP guidelines, and professional certification test guides.
Due to space limitations, experimental results on varying difficulty levels and topic distribution targets are provided in the appendix A. Budgets \(B\in\{0.25C,0.5C,1C\}\) is set relative to the cost \(C\) on the full refinement operation.
Elo updates \(K=64\), \(\eta=400\) with three difficulty bands. Top 3 (\(n\) = 3) distractors are selected for distractor rewriting. GPT‑4o is used as the default backbone LLM for refinement operations if needed. All code and experimental scripts are included in the supplementary material.

\subsection{Experimental Results}
\subsubsection{EQ1: Dataset Quality Improvement.} 

\begin{table}[t]
  \centering \small
  \setlength{\tabcolsep}{1mm}
  \begin{tabular}{c l c c c}
    \toprule
    \textbf{Budget} & \textbf{Strategy}
      & \textbf{Cov. (JSD)$\downarrow$}
      & \textbf{Diff. (JSD)$\downarrow$}
      & \textbf{Dist. (Div.) $\uparrow$} \\
    \midrule
    0.25C 
      & \textbf{RefineLab} & \textbf{0.183} & 0.388 & \textbf{0.243} \\
      & Greedy    & 0.252 & \textbf{0.328} & 0.243 \\
      & Uniform   & 0.250 & 0.418 & 0.213 \\
    \midrule
    0.5C  
      & \textbf{RefineLab} & \textbf{0.063} & \textbf{0.317} & \textbf{0.286} \\
      & Greedy    & 0.190 & 0.342 & 0.286 \\
      & Uniform   & 0.183 & 0.397 & 0.259 \\
    \midrule
    1.0C  
      & \textbf{RefineLab} & \textbf{0.019} & \textbf{0.295} & \textbf{0.331} \\
      & Greedy    & 0.063 & 0.322 & 0.300 \\
      & Uniform   & 0.065 & 0.412 & 0.250 \\
    \bottomrule
  \end{tabular}
  \caption{Cost and quality balance of different strategies} 
  \label{tab:eq2_results}
\end{table}

We evaluate the quality of RefineLab-refined dataset across several dimensions, which are introduced below, along with their respective evaluation metrics. 
1) \textbf{Alignment of Coverage and Difficulty.} We compute the Jensen–Shannon Divergence (JSD) between the empirical distribution \(p\) and the target distribution \(t\), defined as \(\text{JSD}(p \| t) = \frac{1}{2} \mathrm{KL}(p \| M) + \frac{1}{2} \mathrm{KL}(t \| M)\), where \(M = \frac{1}{2}(p + t)\) is the mixture distribution and \(\mathrm{KL}(p \| M) = \sum_i p(i) \log \frac{p(i)}{M(i)}\), where \(i\) indicates the \(i\)-th category (topic or difficulty level) of the distribution. JSD is symmetric and bounded between 0 and 1, with lower values indicating better alignment with the target. 
2)  \textbf{Effectiveness of Distractor Rewriting.} For each multiple-choice question, we evaluate the effectiveness of distractor rewriting, by computing the Shannon entropy of distractor-type proportions \(p_1, \ldots, p_K\) as \(H = -\sum_{k=1}^K p_k \log p_k\), and normalize it by the maximum entropy \(\log K\), where \(K\) indicates total number of mistake types, yielding the normalized entropy \(H_{\text{norm}} = H / \log K\). We report the average of \(H_{\text{norm}}\) across the dataset; higher values indicate greater distractor diversity and improved question discriminability.
3)  \textbf{Successful Correction Ratio.} We measure the correction rate on samples originally flagged as \emph{error} by the Validator, i.e., reporting the proportion that are successfully corrected after refinement based on human annotation. 
4) \textbf{Overall Error Rate.} we draw a
stratified random sample of 400 items from each refined dataset, and report the overall error rate based on human evaluation. 
The human evaluation process involves independent review of each sample by three human annotators with domain expertise. Each annotator assesses whether the QA pair contains any factual, logical, or structural flaws. Annotators are blind to the refinement strategy and base their judgments solely on correctness. The majority vote is used to determine the ground-truth correctness of each sample.

Table~\ref{tab:eq1_results} summarizes the evaluation results. As it shows, RefineLab substantially reduces divergence from the target distributions: coverage JSD decreases by over 90\% and difficulty JSD by over 70\%. 
Additionally, the effectiveness of distractor rewriting is evident, with distractor diversity increasing from 40\% to 90\%, demonstrating that the refined distractors capture a broader and more informative set of misconception patterns. The high correction rates  (over 94.4\%)  further indicate that the Validator successfully corrects the vast majority of detected errors.
Finally, across all datasets, RefineLab yields remarkably low error rates (typically below 4\%), indicating that it reliably produces high-quality, valid QA pairs even under constrained generation.
The additional results in the appendix demonstrate the effectiveness of RefineLab in adjusting difficulty levels and ensuring alignment with various target topic distributions.

\begin{figure}[t]
    \centering
    \includegraphics[width=0.8\linewidth]{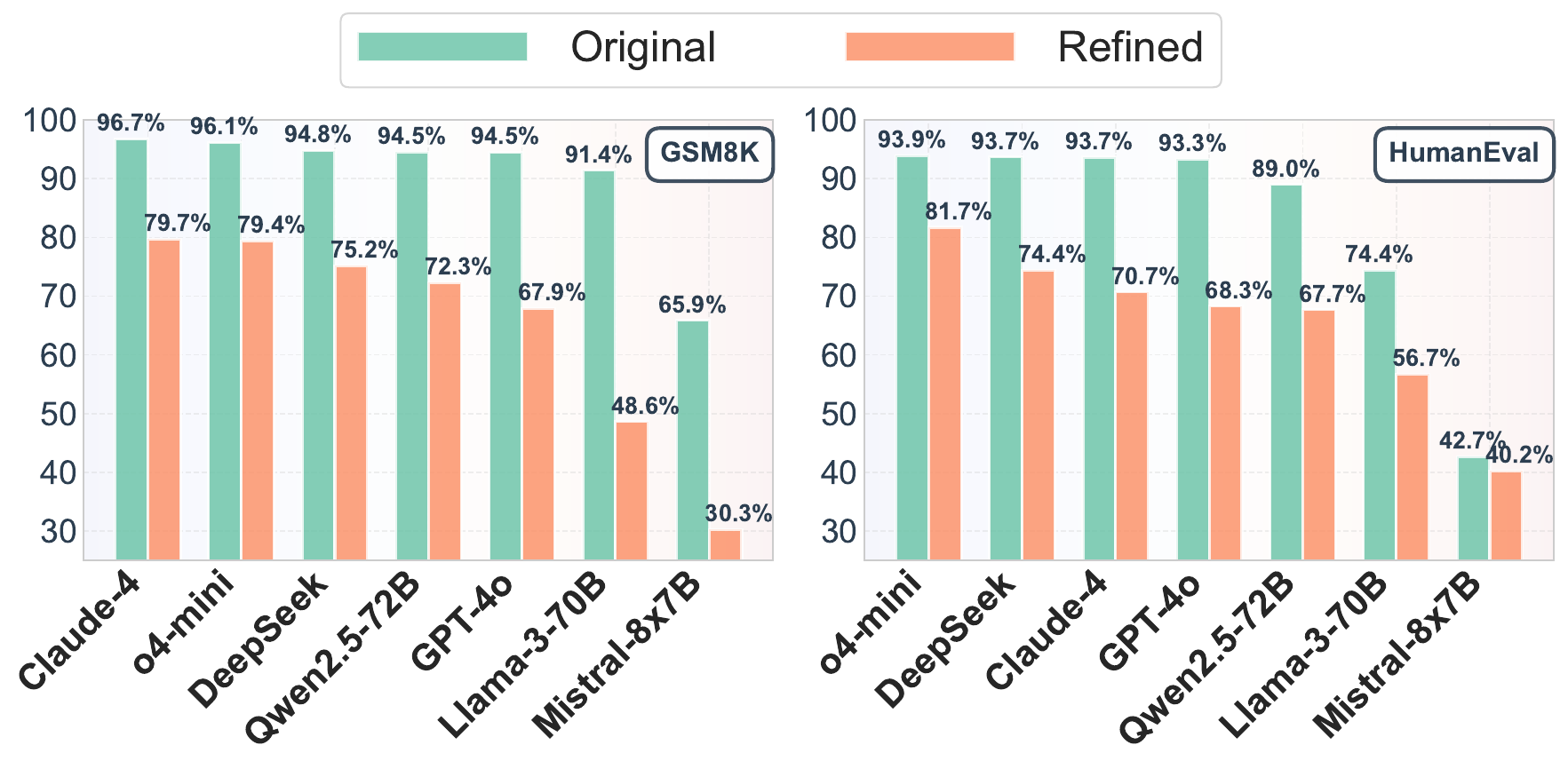}\vspace{-0.1in}
    \caption{Comparison of LLM performance on original and refined datasets.  Left:   GSM8K (original vs. refined). Right:  HumanEval (original vs. refined). }
    \label{fig:evaluation}
\end{figure}


\subsubsection{EQ2: Cost and Quality Balance.} 

To evaluate the effectiveness of our \textit{Assignment Engine} on balancing cost and quality, we conduct experiments on a pilot subset, comprising 10\% of the MMLU dataset, under three budget settings: \(B \in \{0.25\,C,\;0.5\,C,\;1.0\,C\}\), where \(C\) denotes the cost of running all refinement on the pilot subset. We compare three allocation strategies: \textbf{a)} our \textbf{RefineLab}; \textbf{b)} a \textbf{Greedy} approach that iteratively selects the $(q_i,a_i)$–$r_k$ pair with the highest utility–cost ratio, allocating the budget accordingly; \textbf{c)} a \textbf{Uniform} way that evenly splits the budget \(B\) across all operations, with random data selection.

For each strategy and budget, we obtain a refined subset and measure the alignment of coverage and difficulty, as well as the distractor diversity. The total cost per QA sample remains highly affordable,  typically up to \$0.03 per interaction—assuming all refinement operations are applied with GPT-4o as the backbone LLM. As shown in Table~\ref{tab:eq2_results}, RefineLab consistently achieves the best or near-best performance across all metrics, particularly under tight budgets where intelligent allocation is critical. For instance, at \(B = 0.25\,C\), RefineLab achieves significantly lower topic coverage JSD than Greedy, while maintaining strong difficulty alignment. Greedy occasionally performs well on isolated metrics (e.g., difficulty at low budget), but lacks global coordination. As the budget increases, all strategies improve, but RefineLab maintains a clear lead, demonstrating its ability to coordinate multi-dimensional quality improvements more effectively.

\begin{figure}[t]
    \centering
    \includegraphics[width=0.8\linewidth]{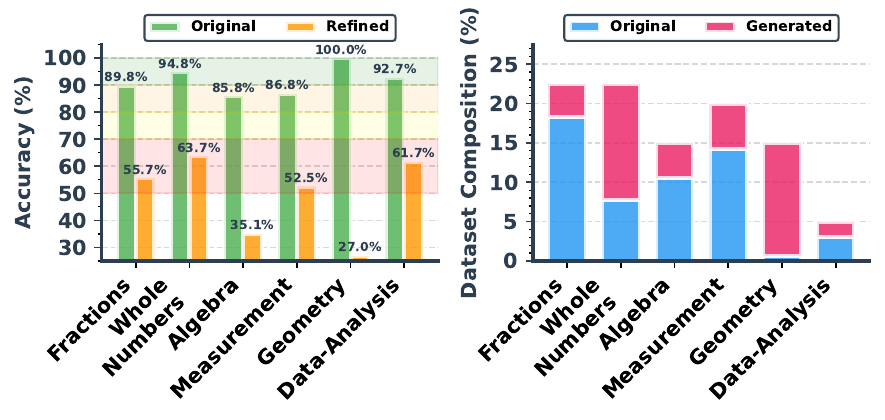}\vspace{-0.1in}
    \caption{Performance breakdown on original and refined GSM8K (left), and refined dataset composition (right).}
    \label{fig:analysis}
\end{figure}

\subsubsection{EQ3: Impact of Refined Datasets on LLM Evaluation.}

We evaluate how RefineLab-refined datasets can affect LLM evaluation by investigating whether \textbf{a) model \textbf{performances} can be better  distinguished}; and \textbf{b) the refined dataset yields consistent model rankings} under different prompting strategies. We employ 7 popular LLMs here: o4-mini \citep{o4-mini}, GPT-4o \citep{gpt4o-hello}, Claude-4-Sonnet \citep{claude-sonnet-4}, DeepSeek-V3 \citep{deepseekai2025deepseekv3technicalreport}, Llama-3-70B-Instruct \citep{llama3modelcard}, Qwen2.5-72B-Instruct \citep{qwen2025qwen25technicalreport}, and Mistral-8x7B-Instruct-v0.1 \citep{jiang2024mixtralexperts}.

\textbf{a) Improved Performance Distinction}.
Figure~\ref{fig:evaluation} presents a comparison of LLM performance on both original and refined datasets. On the left, it shows the performance of various LLMs on the original GSM8K dataset versus the refined version. On the right, a similar comparison is made for HumanEval, illustrating how the refined dataset affects model performance. On GSM8K, the original dataset shows that top-tier LLMs (the top 5) are clustered within a narrow 2.2\% range, masking the true performance gaps. After refinement, the performance spread expands to 11.8\%, with the gap between o4-mini and Mistral-8x7B-Instruct-v0.1 increasing from 30.8\% to 49.4\%. Similarly, on HumanEval, the original dataset has a narrow 0.6\% spread among the top models, which grows to 13.4\% after refinement, with weaker models showing more noticeable relative shifts.
To better understand this widening gap, we analyze Llama-3-70B-Instruct performance by topic in refined GSM8K (Figure~\ref{fig:analysis}). The refined dataset increases representation in categories such as \textit{Geometry}, which are underrepresented in the original, by generating \textit{hard} examples. It corresponds to the largest performance drops, with accuracies decreasing by over 60\%. In contrast, accuracy drops less on topics such as \textit{Whole Numbers}. This selective degradation reveals that RefineLab surfaces previously obscured model weaknesses by rebalancing coverage and injecting difficulty into subdomains, thereby enhancing the dataset’s ability to distinguish true capability differences among LLMs.

\begin{figure}[t]
    \centering
    \includegraphics[width=0.7\linewidth]{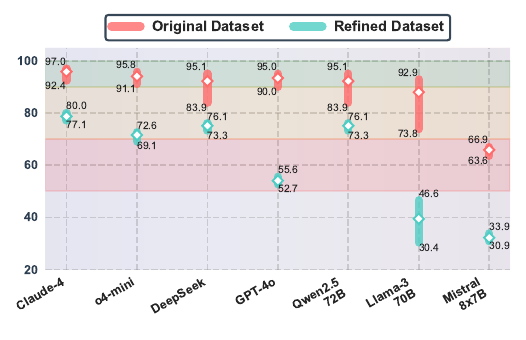} \vspace{-0.15in}
    \caption{Performance variation on GSM8K across prompting strategies, comparing original vs. refined datasets.} 
    \label{fig:stability}
\end{figure}

\begin{figure}[t]
    \centering
    \includegraphics[width=0.6\linewidth]{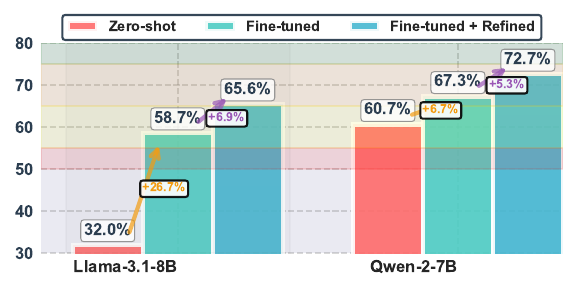}
    \vspace{-0.1in}
    \caption{Performance improvements from fine-tuning with RefineLab augmentations compared to the original dataset.}
    \label{fig:fine-tune}
\end{figure}

\textbf{b) Consistent Model Rankings.}
Figure~\ref{fig:stability} plots the variation in model performance across different prompting strategies for GSM8K. 
Specifically, we evaluate on the GSM8K dataset using 6 prompting variants drawn from: \textit{role play prompting}, \textit{chain-of-thought reasoning}, and \textit{self-consistency prompting}. We observe that the average performance range across LLMs shrinks from 8.4\% on the original data to 4.9\% after refinement. This indicates that the refined dataset enhances consistency in model rankings.

\subsubsection{EQ4: Refined Datasets for LLM Fine-Tuning.}

To assess RefineLab as a data augmentation tool, we fine‑tune Llama-3.1-8B and Qwen-2-7B using 1200 GSM8K original and RefineLab‑refined data, and evaluate on a 150‑item test set, combining both original and refined samples. Figure~\ref{fig:fine-tune} shows that RefineLab augmentations produce improvements compared with the original dataset. Llama-3.1-8B improves from 58.7\% to 65.6\%, while Qwen-2-7B rises from 67.3\% to 72.7\%, confirming consistent benefits across architectures.

\section{Conclusion}
We introduce RefineLab, an LLM-driven, budget‑aware framework designed to refine QA datasets by optimizing key quality dimensions under token-level cost constraints. RefineLab integrates refinement operations, including coverage alignment and difficulty calibration, and formulates the operation assignment as an ILP problem to enable efficient token-budget allocation. Experiments show that RefineLab markedly reduces distributional gaps, achieves near‑zero label errors, enhances LLMs evaluation, and boosts fine‑tuning efficiency. Future work will extend RefineLab to handle multi-turn and open-ended QA format, where refinement must account for longer contexts and dialogue coherence.

\newpage
\bibliography{aaai2026}
\bibliographystyle{plainnat}

\appendix

\newpage
\section{Additional Experimental Results}
\label{app:data_example}

This section provides an extended analysis of RefineLab’s performance under varying refinement targets. We focus on the generality and adaptability of the \textit{Coverage Refiner} and \textit{Difficulty Calibrator} under alternate topic and difficulty distributions.

\subsection{Topic Coverage Alignment under Varying Curricular Targets}

We evaluate RefineLab's coverage alignment performance using three distinct chemistry curricula as target topic distributions for the College Chemistry subset of MMLU: \textbf{Physical Chemistry Curriculum:} Emphasizes physical chemistry, thermodynamics, and molecular quantum mechanics; \textbf{General Chemistry:} Focuses on atomic theory, stoichiometry, bonding, and equilibrium; \textbf{Organic and Inorganic Track:} Balances organic reaction mechanisms, spectroscopy, and coordination chemistry

These targets were converted into topic weight distributions, which were then used as refinement goals for \textit{Coverage Refiner}.

\begin{table}[h]
\centering
\begin{tabular}{lccc}
\toprule
\textbf{Target Curriculum} & \textbf{Original JSD} & \textbf{Refined JSD} & \textbf{JSD Reduction (\%)} \\
\midrule
Physical Chemistry Curriculum              & 0.223 & 0.012 & 94.6\% \\
General Chemistry & 0.194 & 0.009 & 95.4\% \\
Organic/Inorganic & 0.249 & 0.015 & 94.0\% \\
\bottomrule
\end{tabular}
\caption{Coverage alignment under different chemistry curriculum targets. Lower JSD indicates better alignment.}
\label{tab:topic_targets}
\end{table}

As shown in Table~\ref{tab:topic_targets}, RefineLab consistently achieves high alignment across different curricular goals, reducing JSD by over 94\% in all cases. The strongest performance is observed for the General Chemistry aligned target, likely due to its overlap with baseline topic frequencies. The Organic/Inorganic track, with heavier weighting on less common subfields (e.g., NMR, coordination compounds), posed a slightly greater challenge, yet still yielded a 94.0\% reduction. These results highlight RefineLab's ability to flexibly support alignment for diverse curricular standards.

\subsection{Difficulty Calibration under Custom Profiles}

We evaluate three difficulty target profiles using a 1K-sample subset of GSM8K: \textbf{Stress-test:} 0\% easy, 40\% medium, 60\% hard; \textbf{Progressive mastery:} 40\% easy, 40\% medium, 20\% hard; \textbf{Balanced:} 33\% for each level

\begin{table}[h]
\centering
\begin{tabular}{lccc}
\toprule
\textbf{Target Profile} & \textbf{Original JSD} & \textbf{Refined JSD} & \textbf{JSD Reduction (\%)} \\
\midrule
Progressive mastery        & 0.143 & 0.005 & 96.5\% \\
Stress-test & 0.115 & 0.008 & 93.0\% \\
Balanced      & 0.120 & 0.010 & 91.7\% \\
\bottomrule
\end{tabular}
\caption{Difficulty alignment across different target profiles. Lower JSD indicates better calibration.}
\label{tab:difficulty_targets}
\end{table}

As shown in Table~\ref{tab:difficulty_targets}, RefineLab consistently achieves strong difficulty alignment, reducing JSD by over 91\% across all target profiles. The most precise calibration is observed in the \emph{progressive mastery} setting, suggesting that RefineLab is particularly effective when tasked with generating a mixture of difficulty levels—especially when ample medium-difficulty seeds are available to scaffold both upward and downward refinement. These results highlight RefineLab’s ability to support nuanced difficulty profiling for curriculum-aligned generation and robustness evaluation.

\clearpage

\section{Prompt Template}
\label{app:prompt_template}

\begin{figure*}[h]
\begin{tcolorbox}[
  enhanced,
  colframe=brown!75!black,
  colback=white,
  coltitle=white,
  colbacktitle=brown!75!black,
  width=\linewidth,
  arc=2mm,
  auto outer arc,
  boxrule=0.5pt,
  left=10pt,
  right=10pt,
  drop shadow={black!50!white},
  top=10pt,
  bottom=10pt,
  title=\textbf{Expansion (\(r_2\)) prompt},
  fonttitle=\bfseries,
  title code={\node[rounded corners, fill=blue!75!black, draw=none, text=white] at (frame.title) {\textbf{Coverage Refiner}};},
  attach boxed title to top center={yshift=-2mm},
  boxed title style={sharp corners, size=small},
]

\small
You are an expert educational content creator specializing in \texttt{\{domain\}}. Your task is to generate a high-quality multiple-choice question that tests knowledge in the topic: \texttt{\{topic\}}. This question is intended to fill a topic coverage gap.

\textbf{Input QA:}

\texttt{\{(q\_example, a\_example)\}}

\textbf{Instructions:}
\begin{itemize}
    \item Focus on the \textbf{same topic} as the example.
    \item Follow a similar style and structure.
    \item Include exactly 4 plausible answer choices.
    \item Only one correct answer.
    \item Ensure clarity, factual accuracy, and pedagogical value.
    \item Avoid reusing wording or distractors from the example.
    \item Avoid simply rephrasing the question from the example.
\end{itemize}

\textbf{Output format:}
\begin{verbatim}
  "question": "...",
  "choices": ["A", "B", "C", "D"],
  "answer": 1,
  "correct_choice": "Choice B"
\end{verbatim}

\end{tcolorbox}
\caption{Expansion (\(r_2\)) prompt used in the Coverage Refiner module to generate a new QA item aligned with the topic of the input example.}
\label{fig:example_instruction_decomposition}
\end{figure*}

\begin{figure*}[h]
\begin{tcolorbox}[
  enhanced,
  colframe=brown!75!black,
  colback=white,
  coltitle=white,
  colbacktitle=brown!75!black,
  width=\linewidth,
  arc=2mm,
  auto outer arc,
  boxrule=0.5pt,
  left=10pt,
  right=10pt,
  drop shadow={black!50!white},
  top=10pt,
  bottom=10pt,
  title=\textbf{Generation (\(r_4\)) prompt},
  fonttitle=\bfseries,
  title code={\node[rounded corners, fill=blue!75!black, draw=none, text=white] at (frame.title) {\textbf{Difficulty Calibrator}};},
  attach boxed title to top center={yshift=-2mm},
  boxed title style={sharp corners, size=small},
]

\small
You are an expert educational content creator. Your task is to \textbf{rewrite the following multiple-choice question} so that it matches the specified difficulty level: \texttt{\{target\_difficulty\}}.

\textbf{Input QA:}

\texttt{\{(q\_example, a\_example)\}}

\textbf{Instructions:}
\begin{itemize}
  \item Maintain the same topic and general subject as the original.
  \item Rewrite the question and its options to match \texttt{\{target\_difficulty\}}:
    \begin{itemize}
      \item For \texttt{easy}: simplify language, reduce reasoning steps, remove traps.
      \item For \texttt{medium}: balance language complexity and reasoning steps.
      \item For \texttt{hard}: increase reasoning depth, add subtle distractors, require multi-step logic.
    \end{itemize}
  \item Avoid copying phrases directly from the original.
\end{itemize}

\textbf{Ouput format:}
\begin{verbatim}
  "question": "...",
  "choices": ["A", "B", "C", "D"],
  "answer": 1,
  "correct_choice": "Choice B"
\end{verbatim}

\end{tcolorbox}
\caption{Generation (\(r_4\)) prompt for generating an input QA pair with a target difficulty level.}
\label{fig:prompt_difficulty_rewriting}
\end{figure*}

\begin{figure*}[h]
\begin{tcolorbox}[
  enhanced,
  colframe=brown!75!black,
  colback=white,
  coltitle=white,
  colbacktitle=brown!75!black,
  width=\linewidth,
  arc=2mm,
  auto outer arc,
  boxrule=0.5pt,
  left=10pt,
  right=10pt,
  drop shadow={black!50!white},
  top=10pt,
  bottom=10pt,
  title=\textbf{Pairwise Difficulty Comparison Prompt},
  fonttitle=\bfseries,
  title code={\node[rounded corners, fill=blue!75!black, draw=none, text=white] at (frame.title) {\textbf{Elo Rating Module}};},
  attach boxed title to top center={yshift=-2mm},
  boxed title style={sharp corners, size=small},
]

\small
Compare the difficulty of the two questions below and determine which one is more difficult.

---

\textbf{Question A (New Question):}  
\texttt{\{new\_question\}}

\textbf{Question B (Reference Question — \{seed\_difficulty\} difficulty):}  
\texttt{\{seed\_question\}}

---

\textbf{Scoring Guide (Output = Number from 0.0 to 1.0):}
\begin{itemize}
    \item 0.0–0.3: A is \textbf{much easier} than B
    \item 0.3–0.4: A is \textbf{somewhat easier} than B
    \item 0.4–0.45: A is \textbf{slightly easier} than B
    \item 0.45–0.55: A and B are \textbf{similar in difficulty}
    \item 0.55–0.6: A is \textbf{slightly harder} than B
    \item 0.6–0.7: A is \textbf{somewhat harder} than B
    \item 0.7–1.0: A is \textbf{much harder} than B
\end{itemize}

\textbf{Factors to Consider:}
\begin{itemize}
    \item Conceptual complexity
    \item Number of reasoning steps
    \item Required background knowledge
    \item Cognitive load and mental effort
    \item Time and skills required to solve
\end{itemize}

\textbf{Instructions:}
\begin{enumerate}
    \item Analyze both questions carefully
    \item Consider the reference difficulty level: \texttt{\{seed\_difficulty\}}
    \item Respond with a number between 0.0 and 1.0 indicating relative difficulty
    \item Use the full scale — be precise
\end{enumerate}

\textbf{Expected Output:}  
A single number between \texttt{0.0} and \texttt{1.0}.  
\textit{No explanation. No formatting. Just the number.}

\end{tcolorbox}
\caption{Pairwise comparison prompt used for Elo-based difficulty calibration. The LLM compares a newly generated question to a reference seed to infer relative difficulty.}
\label{fig:prompt_pairwise_difficulty}
\end{figure*}

\begin{figure*}[h]
\begin{tcolorbox}[
  enhanced,
  colframe=brown!75!black,
  colback=white,
  coltitle=white,
  colbacktitle=brown!75!black,
  width=\linewidth,
  arc=2mm,
  auto outer arc,
  boxrule=0.5pt,
  left=10pt,
  right=10pt,
  drop shadow={black!50!white},
  top=10pt,
  bottom=10pt,
  title=\textbf{Mistake Type Mining Prompt (\(r_5\))},
  fonttitle=\bfseries,
  title code={\node[rounded corners, fill=blue!75!black, draw=none, text=white] at (frame.title) {\textbf{Distractor Calibrator}};},
  attach boxed title to top center={yshift=-2mm},
  boxed title style={sharp corners, size=small},
]

\small
You are an expert educator in \texttt{\{subject\}}. Your task is to analyze the following question and generate exactly \texttt{\{num\_choices\}} distinct types of mistakes that students might make when attempting to answer it.

---

\textbf{Question:} \texttt{\{question\}}  
\textbf{Correct Answer:} \texttt{\{correct\_answer\}}  
\textbf{Subject:} \texttt{\{subject\}}

---

Each mistake type should:
\begin{itemize}
  \item Be specific to this question and subject
  \item Represent a distinct conceptual or procedural error
  \item Be plausible for partially knowledgeable students
  \item Lead to a wrong answer choice (i.e., be a distractor rationale)
\end{itemize}

\textbf{Examples of mistake types:}
\begin{itemize}
  \item Computational error (e.g., sign error, miscalculation)
  \item Conceptual confusion (e.g., conflating similar terms)
  \item Procedural flaw (e.g., applying wrong formula or method)
  \item Memory lapse (e.g., forgetting a definition or fact)
  \item Logical misstep (e.g., faulty deduction, overgeneralization)
\end{itemize}

\textbf{Respond ONLY with a JSON object like:}
\begin{verbatim}
{
  "mistake_types": [
    {
      "type": "Brief label",
      "description": "Explanation of mistake type 1"
    },
    {
      "type": "Brief label",
      "description": "Explanation of mistake type 2"
    },
    ...
  ]
}
\end{verbatim}

\end{tcolorbox}
\caption{Mistake type mining prompt used in the Distractor Calibrator (\(r_5\)) to elicit plausible misconception patterns for constructing high-quality distractors.}
\label{fig:prompt_mistake_types}
\end{figure*}

\begin{figure*}[h]
\begin{tcolorbox}[
  enhanced,
  colframe=brown!75!black,
  colback=white,
  coltitle=white,
  colbacktitle=brown!75!black,
  width=\linewidth,
  arc=2mm,
  auto outer arc,
  boxrule=0.5pt,
  left=10pt,
  right=10pt,
  drop shadow={black!50!white},
  top=10pt,
  bottom=10pt,
  title=\textbf{Distractor Rewriting Prompt (\(r_5\))},
  fonttitle=\bfseries,
  title code={\node[rounded corners, fill=blue!75!black, draw=none, text=white] at (frame.title) {\textbf{Distractor Calibrator}};},
  attach boxed title to top center={yshift=-2mm},
  boxed title style={sharp corners, size=small},
]

\small
You are an expert educator and assessment designer in \texttt{\{subject\}}. Your task is to \textbf{rewrite all distractors} (i.e., incorrect answer choices) for the following multiple-choice question to improve their pedagogical quality and diagnostic power.

---

\textbf{Question:} \texttt{\{question\}}  
\textbf{Correct Answer:} \texttt{\{correct\_answer\}}  
\textbf{Subject:} \texttt{\{subject\}}

\vspace{0.5em}
\textbf{Original Choices:}
\begin{itemize}
  \item A) \texttt{\{original\_choices[0]\}}
  \item B) \texttt{\{original\_choices[1]\}}
  \item C) \texttt{\{original\_choices[2]\}}
  \item D) \texttt{\{original\_choices[3]\}}
\end{itemize}

\vspace{0.5em}
\textbf{Target Mistake Types (one per distractor):}
\begin{itemize}
  \item[1.] \texttt{\{type\_1\}: \{description\_1\}}
  \item[2.] \texttt{\{type\_2\}: \{description\_2\}}
  \item[3.] \texttt{\{type\_3\}: \{description\_3\}}
\end{itemize}

---

\textbf{Rewrite Instructions:}
\begin{itemize}
  \item Keep the correct answer unchanged: \texttt{\{correct\_answer\}}
  \item Replace all distractors so that:
    \begin{itemize}
      \item Each targets a specific mistake type listed above
      \item Each is plausible to a partially knowledgeable student
      \item Each is clearly incorrect to an expert
      \item All are similar in length and tone to the original
    \end{itemize}
\end{itemize}

\end{tcolorbox}
\caption{Distractor rewriting prompt used in the Distractor Calibrator (\(r_5\)) to rewrite distractors targeting specific misconception patterns, enhancing question difficulty and diagnostic fidelity.}
\label{fig:prompt_distractor_rewriting}
\end{figure*}

\begin{figure*}[h]
\begin{tcolorbox}[
  enhanced,
  colframe=brown!75!black,
  colback=white,
  coltitle=white,
  colbacktitle=brown!75!black,
  width=\linewidth,
  arc=2mm,
  auto outer arc,
  boxrule=0.5pt,
  left=10pt,
  right=10pt,
  drop shadow={black!50!white},
  top=10pt,
  bottom=10pt,
  title=\textbf{Code-based validation prompt},
  fonttitle=\bfseries,
  title code={\node[rounded corners, fill=blue!75!black, draw=none, text=white] at (frame.title) {\textbf{r₆a}};},
  attach boxed title to top center={yshift=-2mm},
  boxed title style={sharp corners, size=small},
]

\texttt{You are a mathematical reasoning expert. For this math problem, generate Python code that solves it step by step.}

\medskip
\texttt{Problem: \{question\}}

\texttt{Answer choices:} \\
\texttt{A) \{choices[0]\}} \\
\texttt{B) \{choices[1]\}} \\
\texttt{C) \{choices[2]\}} \\
\texttt{D) \{choices[3]\}}

\medskip
\texttt{Claimed correct answer: \{correct\_answer\_letter\}) \{correct\_answer\_text\}}

\medskip
\texttt{Generate Python code that:}
\begin{itemize}
  \item Solves the problem step by step
  \item Shows all calculations
  \item Returns the final answer
  \item Includes print statements explaining each step
\end{itemize}

\texttt{The code should be self-contained and executable. Use only standard Python libraries (math, numpy if needed).}

\medskip
\texttt{Respond with ONLY the Python code, no explanations before or after:}

\texttt{\textbackslash\textbackslash\`python} \\
\texttt{\# Your code here} \\
\texttt{\textbackslash\textbackslash\`}
\end{tcolorbox}
\caption{Code-based validation prompt used in math correctness verification.}
\label{fig:code_validation_prompt}
\end{figure*}

\begin{figure*}[h]
\begin{tcolorbox}[
  enhanced,
  colframe=brown!75!black,
  colback=white,
  coltitle=white,
  colbacktitle=brown!75!black,
  width=\linewidth,
  arc=2mm,
  auto outer arc,
  boxrule=0.5pt,
  left=10pt,
  right=10pt,
  drop shadow={black!50!white},
  top=10pt,
  bottom=10pt,
  title=\textbf{RAG-based validation prompt},
  fonttitle=\bfseries,
  title code={\node[rounded corners, fill=blue!75!black, draw=none, text=white] at (frame.title) {\textbf{r₆b}};},
  attach boxed title to top center={yshift=-2mm},
  boxed title style={sharp corners, size=small},
]

\texttt{You are an expert in \{subject\}. Answer this multiple choice question using the provided Wikipedia evidence and your knowledge.}

\medskip
\texttt{Question: \{question\}}

\texttt{Answer choices:} \\
\texttt{A) \{choices[0]\}} \\
\texttt{B) \{choices[1]\}} \\
\texttt{C) \{choices[2]\}} \\
\texttt{D) \{choices[3]\}}

\medskip
\texttt{Wikipedia Evidence: \{summaries\_text\}}

\texttt{Original claimed answer: \{correct\_answer\_letter\}) \{choices[correct\_answer\_index]\}}

\medskip
\texttt{Instructions:}
\begin{itemize}
  \item Use BOTH your expert knowledge and the Wikipedia evidence to determine the correct answer
  \item If Wikipedia evidence is insufficient, rely on your comprehensive knowledge
  \item Consider the context, historical facts, scientific principles, or other relevant information
  \item Choose the answer choice that is most accurate and well-supported
\end{itemize}

\texttt{Think through this step by step:}
\begin{enumerate}
  \item What does the question ask?
  \item What relevant information do you know about this topic?
  \item What does the Wikipedia evidence tell us?
  \item Which answer choice is best supported?
\end{enumerate}

\medskip
\texttt{Respond with ONLY a JSON object:}
\begin{verbatim}
{
  "reasoning": "Your step-by-step reasoning using the evidence",
  "selected_answer": "A, B, C, or D - your chosen answer",
  "original_is_truthful": true/false,
  "confidence": 0.90
}
\end{verbatim}
\end{tcolorbox}
\caption{RAG-based validation prompt for factual correctness checking using Wikipedia.}
\label{fig:rag_validation_prompt}
\end{figure*}

\clearpage

\section{Human Evaluation Details For Refined Data Quality}
\label{app:human_eval_response}

\textbf{Evaluation Guideline.} The human evaluation protocol is summarized in \autoref{fig:human_eval_guideline}. To ensure domain relevance and annotation consistency, the evaluation was conducted over a stratified random sample of 400 refined QA pairs. Each item was independently reviewed by three Ph.D. students with subject-matter expertise in related domain education. Annotators were instructed to judge whether the \texttt{question–answer pair} is (i) factually and logically correct, and (ii) pedagogically appropriate for the stated domain and difficulty level. Each sample was labeled as \textbf{1 (Pass)}, \textbf{0 (Fail)}, or \textbf{N/A (Uncertain)}. For final analysis, samples labeled as \textbf{N/A} were excluded to ensure robust agreement and statistical validity.

\begin{figure*}
\begin{tcolorbox}[definitionbox, title=Human Evaluation Guideline]

\textbf{Objective}

The purpose of this task is to evaluate the quality of refined multiple-choice QA items. Specifically, you should determine whether the \textbf{question and its provided answer} are both \textbf{factually accurate} and \textbf{pedagogically appropriate}.

\vspace{1mm}
\textbf{Files and Format}

Each item will be presented as a dictionary with fields:
\begin{itemize}[leftmargin=5mm]
    \item \texttt{question}: The multiple-choice question text.
    \item \texttt{choices}: The list of four answer options.
    \item \texttt{answer}: The index of the correct answer (0--3).
    \item \texttt{correct\_choice}: The text of the correct answer.
\end{itemize}

Focus your evaluation on whether the QA pair is correct and meaningful in its domain.

\vspace{1mm}
\textbf{Evaluation Criteria}

Assign one of the following labels to each QA pair:

\begin{itemize}[leftmargin=5mm]
    \item \textbf{1 (Pass)}: The correct answer is valid and unambiguous. The question is factually correct, logically coherent, and domain-appropriate.
    \item \textbf{0 (Fail)}: The correct answer is factually incorrect, logically flawed, ambiguous, or the question itself is problematic or misleading.
    \item \textbf{N/A (Uncertain)}: You are unable to make a confident judgment due to unclear content, ambiguous phrasing, or insufficient domain familiarity.
\end{itemize}

\textbf{Important:} Focus strictly on factual and structural validity — do not judge based on stylistic preferences or subjective difficulty unless it causes confusion or invalidation.

\end{tcolorbox}
\caption{Human evaluation guideline for refined QA item quality.}
\label{fig:human_eval_guideline}
\end{figure*}

\clearpage

\section{Broader impacts}
\label{app:broader_impact}

RefineLab represents a critical advancement in data-centric AI, enabling scalable, reliable, and cost-effective refinement of QA datasets for LLM evaluation. By automating the improvement of dataset quality across dimensions like topic coverage, difficulty balance, and factual correctness, RefineLab helps bridge the gap between academic benchmarks and real-world task requirements. This has meaningful implications for both research communities and downstream applications that rely on high-quality evaluation protocols.

\textbf{RefineLab can improve LLM benchmarking and model alignment.}
Current LLM benchmarks often suffer from structural biases and outdated distributions, leading to misleading model comparisons. RefineLab introduces a principled framework for dataset curation that enables controlled, reproducible modifications aligned with pedagogical standards or evaluation goals. This allows model developers to identify failure cases, fine-tune models with targeted augmentations, and design more rigorous evaluation pipelines—ultimately accelerating safe and effective model alignment.

\textbf{RefineLab can empower domain-specific LLMs and agent systems.}
By supporting structured data refinement under constraints, RefineLab provides a data backbone for domain-adapted LLMs and agent systems. Refined datasets can be used to train and evaluate agents in complex settings such as legal reasoning, biomedical QA, or scientific tutoring, where dataset quality critically impacts performance. As foundation models are increasingly deployed in specialized domains, frameworks like RefineLab will be essential for ensuring data integrity, transparency, and evaluation robustness.

\section{Future Work}
\label{app:future_work}
While RefineLab demonstrates strong performance on textual multiple-choice QA datasets, several extensions remain open for future exploration.

\textbf{Refining alternative data formats.}
Future work can extend RefineLab beyond multiple-choice questions to include other QA formats such as open-ended answers, multi-turn dialogues, and chain-of-thought reasoning traces. These formats introduce new refinement challenges—such as coherence, completeness, and reasoning validity—that require adapted refinement operators and quality estimation techniques.

\textbf{Supporting multi-modal datasets.}
Many high-impact evaluation settings involve multimodal content, such as image-based science exams or graph-structured biomedical queries. Adapting RefineLab to handle vision–language QA pairs will require incorporating modality-aware refinement strategies and corresponding validators.

\textbf{Incorporating broader data quality metrics.}
Current refinement focuses on topic alignment, difficulty, and correctness. Future work can incorporate additional quality dimensions that influence LLM evaluation reliability—such as linguistic diversity, reasoning transparency, or question ambiguity. These refinements may require new types of LLM-based feedback and multi-objective assignment formulations. 

\end{document}